%% file: double.tex
\documentclass[journal]{IEEEtran}
\input{commands}

\usepackage{booktabs}
\usepackage{xcolor}
\usepackage{colortbl}
\usepackage{amsmath}

\newcommand{\etal}{{\it{et al}.} }

\makeatletter
\newcommand*{\rom}[1]{\expandafter\@slowromancap\romannumeral #1@}
\makeatother

\begin{document}
%\title{Toward Compressed Image Generation using \\Generative Adversarial Networks}
\title{Toward Joint Image Generation and Compression using Generative Adversarial Networks}
\author{Byeongkeun~Kang,~Subarna~Tripathi,~and~Truong~Q.~Nguyen,~\IEEEmembership{Fellow,~IEEE}
\thanks{B. Kang, S. Tripathi, and T. Q. Nguyen are with the Department of Electrical and Computer Engineering, University of California, San Diego, CA 92093 USA (e-mail: bkkang@ucsd.edu, stripath@ucsd.edu, tqn001@eng.ucsd.edu).}
\thanks{This work is supported in part by NSF grant IIS-1522125.}
}
\maketitle

%\begin{IEEEkeywords}
%Motion trajectories, articulated motion, mesh evolution.
%\end{IEEEkeywords}
\IEEEpeerreviewmaketitle

\input{sections/abstract}

\input{sections/introduction}

\input{sections/related_works}

\input{sections/method}

\input{sections/results}

\input{sections/conclusions}

\footnotesize
\bibliographystyle{IEEEbib}
\bibliography{egbib}

\end{document}

%% file: commands.tex
\usepackage{subfigure}
\usepackage{cite}
\usepackage[pdftex]{graphicx}
\graphicspath{ {images/} }
\usepackage{amsmath}
\usepackage{algorithmic}
\usepackage{array}
\ifCLASSOPTIONcompsoc
  \usepackage[caption=false,font=normalsize,labelfont=sf,textfont=sf]{subfig}
\else
  \usepackage[caption=false,font=footnotesize]{subfig}
\fi
\usepackage{url}

\usepackage{amssymb}
\usepackage{multirow}
\usepackage{tabu}
\usepackage{pbox}
{\begin{list}               %    with flush left bullets.
    {$\bullet$ \hfill}{
        \setlength{\leftmargin}{\parindent}
        \setlength{\parsep}{0.04\baselineskip}
        \setlength{\itemsep}{0.5\parsep}
        \setlength{\labelwidth}{\leftmargin}
        \setlength{\labelsep}{0em}}
    }
{\end{list}}

\providecommand{\eref}[1]{\eqref{#1}}  % call \eqref from amstex
\providecommand{\cref}[1]{Chapter~\ref{#1}}
\providecommand{\sref}[1]{Section~\ref{#1}}
\providecommand{\fref}[1]{Fig.~\ref{#1}}
\providecommand{\tref}[1]{Table~\ref{#1}}

\providecommand{\PP}{\ensuremath{\mathbb{P}}}

\providecommand{\E}{\ensuremath{\mathbb{E}}}

\providecommand{\abs}[1]{\lvert#1\rvert}

\renewcommand{\vec}[1]{\ensuremath{\boldsymbol{#1}}}
\providecommand{\mat}[1]{\ensuremath{\boldsymbol{#1}}}

\providecommand{\mC}{\mat{C}}

\providecommand{\vm}{\vec{m}}

\providecommand{\vx}{\vec{x}}

\providecommand{\xtilde}{\tilde{\vx}}

\providecommand{\vxh}{\hat{\vx}}

%% file: sections/abstract.tex
\begin{abstract}
In this paper, we present a generative adversarial network framework that generates compressed images instead of synthesizing raw RGB images and compressing them separately. In the real world, most images and videos are stored and transferred in a compressed format to save storage capacity and data transfer bandwidth. However, since typical generative adversarial networks generate raw RGB images, those generated images need to be compressed by a post-processing stage to reduce the data size. Among image compression methods, JPEG has been one of the most commonly used lossy compression methods for still images. Hence, we propose a novel framework that generates JPEG compressed images using generative adversarial networks. The novel generator consists of the proposed locally connected layers, chroma subsampling layers, quantization layers, residual blocks, and convolution layers. The locally connected layer is proposed to enable block-based operations. We also discuss training strategies for the proposed architecture including the loss function and the transformation between its generator and its discriminator. The proposed method is evaluated using the publicly available CIFAR-10 dataset and LSUN bedroom dataset. The results demonstrate that the proposed method is able to generate compressed data with competitive qualities. The proposed method is a promising baseline method for joint image generation and compression using generative adversarial networks.
\end{abstract}

%% file: sections/introduction.tex
\section{Introduction} 
\label{introduction}

Most images and videos exist in a compressed form since data compression saves lots of data storage and network bandwidth and further enables many applications such as real-time video streaming in cell phones. Compression is indeed crucial, considering compressed image and video can be about 10 times and 50 times smaller than raw data, respectively. Nevertheless, typical generative adversarial networks (GAN) focus on generating raw RGB images or videos~\cite{GAN_NIPS2014, dcgan_radford2015unsupervised, WassersteinGAN, NIPS2017_7159, Vondrick, Xiong}. Considering one of the most common usages of GANs is generating large-scale synthetic images/videos for data augmentation, the created images/videos often require compression in a post-processing stage to store the large dataset in a hardware~\cite{Sixt, wang2018pix2pixHD, wang2018vid2vid, FridAda}. Besides, typical GANs are evaluated using the generated raw RGB data although compression is processed to the raw data prior to final applications. Hence, we investigate GAN frameworks that aim to generate compressed data and to evaluate the networks using the generated encoded images.

We focus on the GAN frameworks for compressed image generation since image generation networks~\cite{GAN_NIPS2014, dcgan_radford2015unsupervised, WassersteinGAN, NIPS2017_7159} have been far more studied comparing to video generation networks~\cite{Vondrick, Xiong}. In image compression, JPEG~\cite{ITU1982, ITU1992} has been one of the most commonly used lossy compression methods~\cite{Wallace, Pennebaker, Hudson}. It has been used in digital cameras and utilized for storing and transmitting images. While JPEG has many variants, typical JPEG consists of color space transformation, chroma subsampling, block-based discrete cosine transform (DCT)~\cite{DCT_1974}, quantization, and entropy coding~\cite{Huffman}. In more details, the compression method converts an image in the RGB domain to the image in another color space (YCbCr) that separates the luminance component and the chrominance components. Then, the chrominance components are downsampled. It then applies the 8$\times$8 block-based DCT to both the luminance component and the subsampled chrominance components to represent them in the frequency domain. It discards details of high-frequency information by applying quantization. Lastly, the processed data is stored using entropy coding.

\begin{figure}
\begin{center}
    \includegraphics[width=0.45\textwidth]{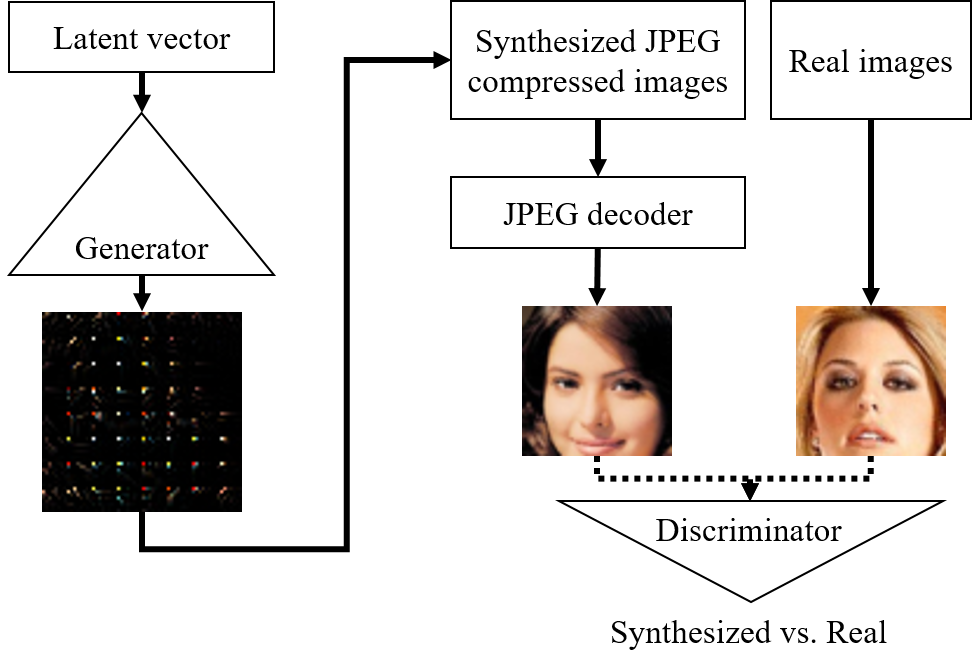} 
\end{center}
   \caption{The proposed framework to generate compressed images. The framework consists of a generator, a discriminator, and a transformer between the generator and the discriminator. The visualized generator output is an intermediate example.}
   % The proposed generator produces compressed data given a randomly selected latent vector. The transformer is employed to make the data from both the generator and the training data in the same domain. Since the generator outputs compressed images and the training data is raw RGB images, the transformer is a decoder. The discriminator takes synthesized images and real images and aims to differentiate them.}
\label{proposed_framework}
\end{figure}

We argue that investigating the frameworks of generating compressed images is important to accomplish the creation of more visually plausible large-scale images that require storing in a compressed domain. Typical GAN frameworks optimize and select the networks' architectures and parameters (weights) based on generated raw RGB images. Accordingly, if we take into account the compression process in the post-processing stage, the choice might not be the optimal decision. Hence, we propose to optimize/determine architectures and parameters by evaluating them using generated images in a compression domain.

We propose a novel framework that generates compressed images using generative adversarial networks as shown in~\fref{proposed_framework}. The framework consists of a generator, a discriminator, and a transformer between the generator and the discriminator. The proposed generator produces compressed data given a randomly selected noise in a latent space. The transformer is applied to make the data from the generator and the training data to be in the same domain. Since the generator outputs compressed data and the training data is raw RGB images, the transformer should convert encoded data to a raw image, so is a decoder. The discriminator takes synthesized images and real images and aims to differentiate them. 

The proposed generator has three paths, one path for the luminance component and the other two paths for the chrominance components. The separate paths are proposed to process any requested chroma subsampling. We also propose the locally connected layer that takes an input of a subregion and outputs for the corresponding subregion. The proposed locally connected layer is able to handle block-based processing in JPEG compression. In summary, the contributions of our work are as follows:

\begin{itemize}
  \item We propose the framework that generates JPEG compressed images using generative adversarial networks.
  \item We propose the locally connected layer to enable block-based operations in JPEG compression. 
  \item We analyze the effects of compression for the proposed method and other methods.
\end{itemize}

%% file: sections/related_works.tex
\section{Related Works} 
\subsection{Generative Adversarial Networks}
Generative adversarial networks were introduced in~\cite{GAN_NIPS2014} where the framework is to estimate generative models by learning two competing networks (generator and discriminator). The generator aims to capture the data distribution by learning the mapping from a known noise space to the data space. The discriminator differentiates between the samples from the training data and those from the generator. These two networks compete since the goal of the generator is making the distribution of generated samples equivalent to that of the training data while the discriminator's objective is discovering the discrepancy between the two distributions.
 
While the work in~\cite{GAN_NIPS2014} employed multilayer perceptrons for both generator and discriminator, deep convolutional GANs (DCGANs) replaced multilayer perceptrons by convolutional neural networks (CNNs) to take the advantage of shared weights, especially for image-related tasks~\cite{dcgan_radford2015unsupervised}. To utilize CNNs in the GAN framework, extensive architectures with the relatively stable property during training, are explored. They examined stability even for models with deeper layers and for networks that generate high-resolution outputs. The analysis includes fractional-stride convolutions, batch normalization, and activation functions. Salimans \etal presented the methods that improve the training of GANs~\cite{Salimans}. The techniques include matching expected activations of training data and those of generated samples, penalizing similar samples in a mini-batch, punishing sudden changes of weights, one-sided label smoothing, and virtual batch normalization.  

Arjovsky \etal presented the advantage of the Earth-Mover (EM) distance (Wasserstein-1) comparing to other popular probability distances and divergences such as the Jensen-Shannon divergence, the Kullback-Leibler divergence, and the Total Variation distance~\cite{WassersteinGAN}. The advantage of the EM distance is that it is continuous everywhere and differentiable almost everywhere when it is applied for a neural network-based generator with a constrained input noise variable. They also showed that the EM distance is a more sensible cost function. Based on these, they proposed Wasserstein GAN that uses a reasonable and efficient approximation of the EM distance. They then showed that the proposed GAN achieves improved stability in training. However, clipping weights for Lipschitz constraint in~\cite{WassersteinGAN} might cause optimization difficulties~\cite{NIPS2017_7159}. Hence, Gulrajani \etal proposed penalizing the gradient norm to enforce Lipschitz constraint instead of clipping~\cite{NIPS2017_7159}. 

Wasserstein GAN trains its discriminator multiple times at each training of its generator so that the framework can train the generator using the more converged discriminator~\cite{WassersteinGAN}. To avoid the expensive multiple updates of the discriminator, Heusel \etal proposed to use the two time-scale update rule (TTUR)~\cite{Prasad} in a Wasserstein GAN framework~\cite{NIPS2017_7240}. Since TTUR enables separate learning rates for the discriminator and the generator, they can train the discriminator faster than the generator by selecting a higher learning rate for the discriminator comparing to that of the generator. It is also proved that TTUR converges to a stationary local Nash equilibrium under mild assumptions. They further experimentally showed that their method outperforms most other state-of-the-art methods. Hence, the proposed framework of this paper is based on \cite{NIPS2017_7240}. 

\input{sections/fig_architecture}

\subsection{Image Compression: JPEG}
JPEG~\cite{ITU1992} has been one of the most commonly used lossy compression methods for still images with continuous tones~\cite{Wallace, Pennebaker, Hudson}. It has been used for digital cameras, photographic images on the World Wide Web, medical images, and many other applications.

In JPEG, discrete cosine transform (DCT) is utilized since it achieves high energy compaction while having low computational complexity. Considering an image contains uncorrelated (various) information, block-based DCT is used so that each block contains correlated data. Using too small block prevents from compressing correlated information. Too large block with uncorrelated pixels increases computational complexity without compression gain. 8$\times$8 block size is selected based on a psychovisual evaluation.

DCT transforms an 8$\times$8 block of an image to 64 amplitudes of 2D cosine functions with various frequencies. Since the sensitivity of a human eye is different for each frequency, quantization is applied differently for each amplitude. The amplitudes for low-frequencies are maintained with high accuracy and those of high-frequencies are quantized using larger quantization value. Quantization is responsible for most of the information loss in JPEG.

After quantization, since most of the non-zero components are for low-frequencies, the amplitudes are encoded in zig-zag order using a value pair. The information is then encoded using Huffman coding considering the statistical distribution of the information~\cite{Huffman}.

While these are the baseline of JPEG, other additional methods and components were also suggested for particular purposes. Also, typical JPEG uses the YCbCr color space and chroma subsampling~\cite{ITU1982}.

\subsection{Other Related Works } 
Our work differs from learning neural networks for image compression such as autoencoders~\cite{comp_autoencoder_17, LSTM_img_comp_16, GRU_resnet_16, Gene_comp_17} that aims to learn image encoders to compress real images. The proposed method and learning encoders differ in two aspects. First, the goal of the proposed method is generating synthetic JPEG compressed data while that of the latter is compressing real images. Second, the proposed method intends to utilize existing standard decoder in general electronic devices while the latter requires a particular decoder to decode the compressed data. 

This paper is also distinct from~\cite{Kaneko2017, MardaniGCVZATHD17} that are about utilizing GANs for postprocessing real data in a frequency domain. The proposed work generates compressed images from random noises in the latent space. 

% \cite{comp_autoencoder_17} proposes a method that optimizes autoencoders for lossy image compression.
% \cite{LSTM_img_comp_16} proposes a method based on convolutional and deconvolutional LSTM recurrent networks for variable-rate image compression.
% \cite{GRU_resnet_16} proposes a recurrent neural network (RNN)-based method for lossy image compression.
% \cite{Gene_comp_17} proposes a method that compresses data using generative models.

% \cite{Kaneko2017} propose a learning-based postfilter to reconstruct the high-fidelity spectral texture in short-term Fourier transform (STFT) spectrograms.
% \cite{MardaniGCVZATHD17} propose a novel compressed sensing framework that utilizes GANs to train a (low-dimensional) manifold of diagnostic-quality MR images from historical patients. The generator is learned to remove the aliasing artifacts learns texture details.

%% file: sections/fig_architecture.tex
\begin{figure*}[!t] \begin{center}
\begin{minipage}{0.98\linewidth}
  \centerline{\includegraphics[width=\textwidth]{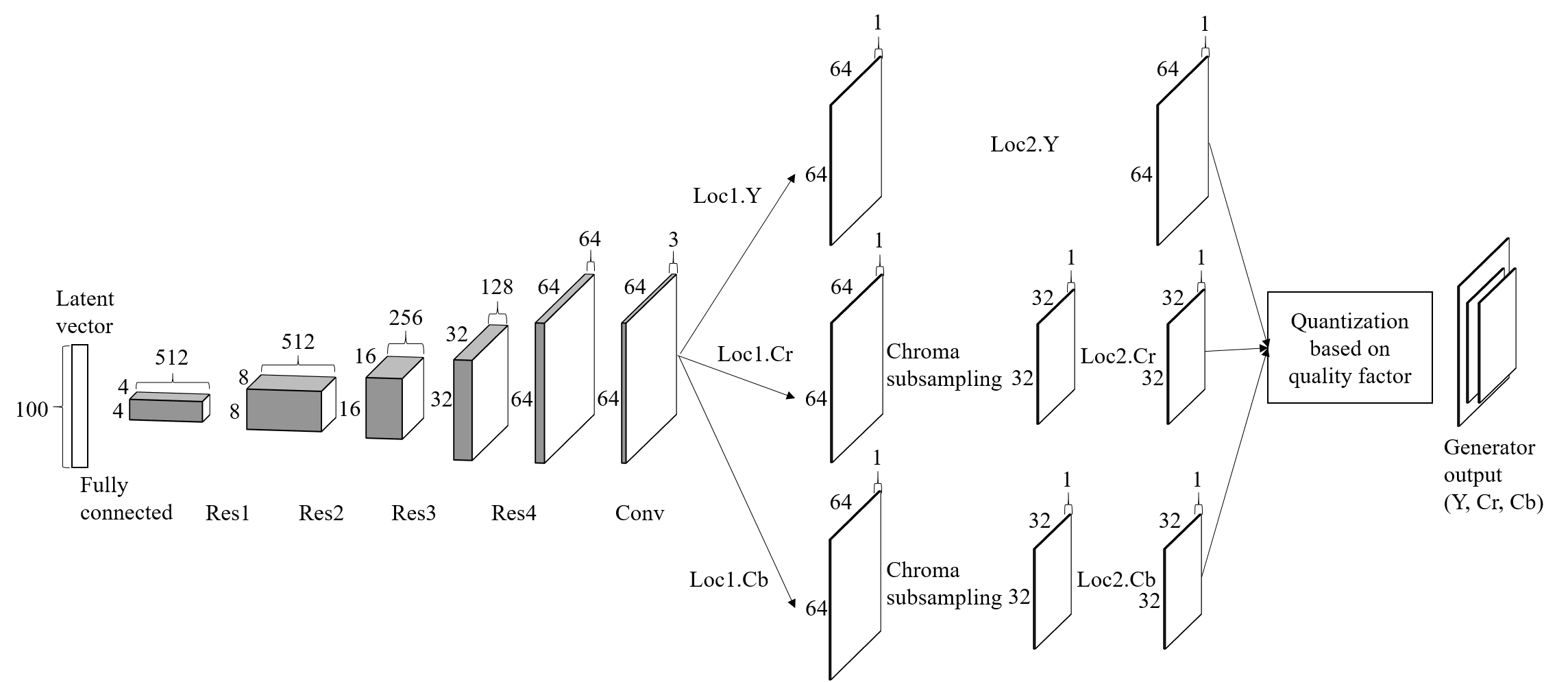}}
  \centerline{\footnotesize (a)}
\end{minipage}
\begin{minipage}{0.49\linewidth}
  \centerline{\includegraphics[width=0.90\textwidth]{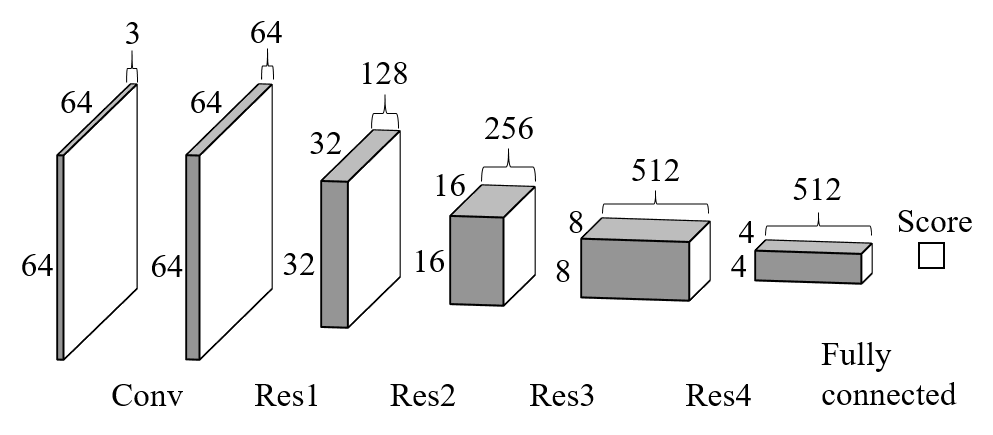}}
  \centerline{\footnotesize (b)}
\end{minipage}
\begin{minipage}{0.49\linewidth}
  \centerline{\includegraphics[width=0.90\textwidth]{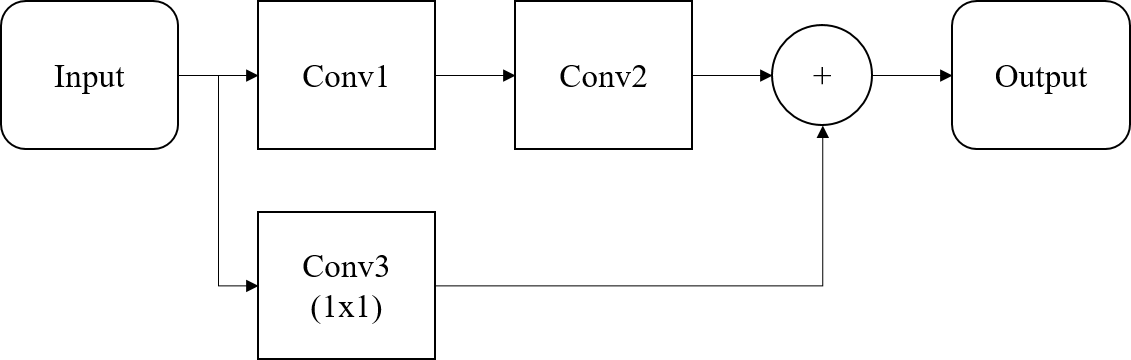}}
  \centerline{\footnotesize (c)}
\end{minipage}
  \caption{The proposed architecture for JPEG compressed image generation. (a) Generator. (b) Discriminator. (c) Residual block. The proposed generator consists of three paths, one for each luminance or chrominance component. The generator employs the proposed locally connected layer to operate block-based processing. The visualized generator considers chroma subsampling ratio of 4:2:0.}
\label{fig:architecture}
\end{center} \end{figure*}

%% file: sections/method.tex
\input{sections/fig_locally_connect}

\section{Proposed Method}  \label{method}
We propose a framework that generates JPEG compressed images using generative adversarial networks. We use the architectures analyzed in the TTUR method~\cite{NIPS2017_7240} as our baseline networks since the method achieves one of the state-of-the-art results. The generator in the baseline architecture consists of one fully connected layer, four residual blocks, and one convolution layer. The discriminator in the architecture consists of one convolution layer, four residual blocks, and one fully connected layer as shown in~\fref{fig:architecture}(b). The residual block consists of two paths (see~\fref{fig:architecture}(c)). One path has two convolution layers with filter-size of 3$\times$3, and the other has only one convolution layer with filter-size of 1$\times$1. All the convolution layers outside of residual blocks have filter-size of 3$\times$3. 

Given the baseline architecture, we propose an architecture and training strategy to generate JPEG compressed images in the framework of generative adversarial networks. We first propose a novel generator in~\sref{architecture}. The proposed generator has three paths, one for each luminance or chrominance component. The proposed generator also has additional layers including the proposed locally connected layers, chroma subsampling layer, and quantization layer. An entropy encoding layer is omitted since the encoding is lossless and located at the last layer. Hence, this exclusion does not affect the results.

We then present the processing between the generator and the discriminator in~\sref{transform}. Since typical generators generate RGB images which are in the same domain with the training images, any additional processing is not required to use the output of the generator for the input to the discriminator. However, since the proposed generator produces JPEG compressed data, the outputs of the generator and the training images are in different domains and cannot be used together for the discriminator. Consequently, we need to either compress the training images or decode the generated JPEG compressed data so that they are in the same domain.

In~\sref{training}, we discuss training strategies for the proposed architecture. Although many studies have been conducted to improve training stability, training GANs for non-typical images is still quite challenging. 

\subsection{Generator} \label{architecture}
We propose a novel generator that generates JPEG compressed images (see~\fref{fig:architecture}). The generator consists of six locally connected layers, two chroma subsampling layers, and a quantization layer in addition to the layers in the baseline generator. The generator has three paths where each path generates one of luminance or chrominance components in the YCbCr representation. The separated paths are required to handle any required chroma subsampling since the resolution of a luminance component and chrominance components are different if chroma subsampling is applied. The locally connected layer is proposed to operate block-based processing. An entropy encoding layer is not applied since the encoding is lossless and at the last layer, which does not impact the results.

The proposed locally connected layer takes an input of a subregion and produces an output for the corresponding subregion (see~\fref{fig:locallyConnect}). The layer is proposed to perform operations comparable to block-based processing. Comparing to a convolution layer, the proposed layer is different since a convolution layer takes an input from a region and outputs to only a single location. The nearby outputs from a convolution layer are produced by using different regions of inputs. In other words, to generate 8$\times$8 outputs using a convolution layer, the layer actually takes inputs from 64 different regions by shifting a filter (weights). The proposed layer is also dissimilar from a fully connected layer since a typical fully connected layer does not share weights while the proposed locally connected layer shares weights between blocks. For all paths in the generator, the first locally connected layer (Loc1) and the second locally connected layer (Loc2) employ the block-size of 1$\times$1 and 8$\times$8, respectively. The block-size of 8$\times$8 is selected considering 8$\times$8 block-based inverse DCT in a JPEG decoder. The block-size of 1$\times$1 is a special case that can be reproduced by a convolution layer.

Chroma subsampling is processed by averaging the amplitudes of the chrominance component of each block and by subsampling from the block to a scalar. In this paper, we investigate the proposed architecture using the popular subsampling ratios, 4:4:4, 4:2:2, and 4:2:0. The 4:4:4 ratio means no subsampling and preserves all the chrominance information. The 4:2:2 mode averages and subsamples with 2:1 ratio for only the horizontal axis. Consequently, the horizontal resolution of the output is half of the input. For the 4:2:0 subsampling, both horizontal and vertical axes are averaged and subsampled with 2:1 ratio. Consequently, each block of 2$\times$2 pixels is turned to a scalar (see~\fref{fig:architecture}(a)).

Forward processing all the layers in the proposed generator before quantization generates amplitudes of 2D cosine functions for luminance and chrominance components. Quantization is then performed using a conventional quantization method in JPEG compression. We employ the conventional method so that the final output is able to be de-quantized by using a typical JPEG decoder. Quantization is performed by dividing amplitudes by quantization matrices and by rounding the quantized amplitudes to an integer~\cite{Pratt, Lohscheller}. The quantization matrices are determined based on a user-selected quality factor and can also be selected in an encoding process. We employ popular quantization matrices that are shown at~\eref{Q_lum} and~\eref{Q_chr}~\cite{Peterson}. The given quantization matrices ($Q_{l,50}$, $Q_{c,50}$) are for the quality factor of 50. The former and the latter matrices are for luminance component ($Q_l$) and chrominance part ($Q_c$). 
\begin{equation}
Q_{l,50}=
\begin{bmatrix}
    16 & 11 & 10 & 16 & 24 & 40 & 51 & 61 \\
    12 & 12 & 14 & 19 & 26 & 58 & 60 & 55 \\
    14 & 13 & 16 & 24 & 40 & 57 & 69 & 56 \\
    14 & 17 & 22 & 29 & 51 & 87 & 80 & 62 \\
    18 & 22 & 37 & 56 & 68 & 109 & 103 & 77 \\
    24 & 35 & 55 & 64 & 81 & 104 & 113 & 92 \\
    49 & 64 & 78 & 87 & 103 & 121 & 120 & 101 \\
    72 & 92 & 95 & 98 & 112 & 100 & 103 & 99 \\
\end{bmatrix}
.
\label{Q_lum}
\end{equation}
\begin{equation}
Q_{c,50}=
\begin{bmatrix}
    17 & 18 & 24 & 47 & 99 & 99 & 99 & 99 \\
    18 & 21 & 26 & 66 & 99 & 99 & 99 & 99 \\
    24 & 26 & 56 & 99 & 99 & 99 & 99 & 99 \\
    47 & 66 & 99 & 99 & 99 & 99 & 99 & 99 \\
    99 & 99 & 99 & 99 & 99 & 99 & 99 & 99 \\
    99 & 99 & 99 & 99 & 99 & 99 & 99 & 99 \\
    99 & 99 & 99 & 99 & 99 & 99 & 99 & 99 \\
    99 & 99 & 99 & 99 & 99 & 99 & 99 & 99 \\
\end{bmatrix}
.
\label{Q_chr}
\end{equation}
Quantization matrix for another quality factor for luminance component is computed as follows:
\begin{equation}
\begin{aligned}
Q_{l,n} =\begin{cases}
              \max{ (1, \lfloor \frac{100-n}{50} Q_{l,50} + 0.5 \rfloor) },  & \text{if } n \geq 50. \\
               \lfloor \frac{50}{n} Q_{l,50} + 0.5  \rfloor, 					& \text{otherwise}. \\
		\end{cases}
\end{aligned}
\end{equation}
where $n \in (0, 100]$ denotes a quality factor. The conversion of the quantization matrix for chrominance is equivalent. 

The architecture in~\fref{fig:architecture} is used for the LSUN bedroom dataset~\cite{yu15lsun} which aims to generate images with the resolution of 64$\times$64. For the CIFAR-10 dataset~\cite{cifar} whose objective resolution is 32$\times$32, the output dimensions of all the layers in both the generator and the discriminator are reduced by half for both x- and y-axes. Also, the number of activations (feature maps) is reduced by half up to the last residual block in both the generator and the discriminator.

\subsection{Transform between Generator and Discriminator} \label{transform}
Since the discriminator takes half of the inputs from the generator and the other half from the training dataset, the two data should be in the same domain (representation). However, the training data set contains real images in the RGB domain and outputs of the proposed generator are JPEG compressed data. Hence, we have to either compress the training data or decode the outputs of the generator so that the two images are in the same domain. 

We examined both alternatives. It turns out that it is better to decode outputs of the generator before providing them for inputs to the discriminator. Our opinion is that convolution layers, which are major elements of the discriminator, are invented for real images which usually have a continuous tone. However, the compressed data contains amplitudes of block-based DCT which vary largely at the boundary of blocks and also in the blocks. Consequently, the discriminator provides inferior-quality gradients to the generator and hinders training a good generator. Hence, the proposed framework has a decoder that takes outputs of the generator and renders inputs to the discriminator during training. We do not need this conversion (decoder) after training since we only utilize the discriminator for training and our goal is generating compressed data. 

Given an output of the generator, we first de-quantize the amplitudes by multiplying them by the corresponding quantization matrices used in the generator. We then apply inverse DCT to transform the amplitudes in the frequency domain to the contents in the 2D color domain. We upsample chrominance components to the same resolution of the luminance component if chroma subsampling is applied in the generator. We then convert the amplitudes in the YCbCr space to the RGB space. Lastly, we clip the amplitudes so that after compensating shifting and scaling, the amplitudes are in the range of [0, 255].

\subsection{Training} \label{training}
As GANs are difficult to train~\cite{GAN_NIPS2014}, many studies have been conducted to improve the stability of training~\cite{Salimans, WassersteinGAN, NIPS2017_7159}. Still, by employing current state-of-the-art training algorithms to our problem, we had difficulty in training the proposed networks. Hence, we propose a novel loss function to train the proposed framework. 

Given a loss function, $L$, the generator $G$ and the discriminator $D$ are trained by playing a minimax game as follows:
\begin{equation}
\begin{aligned}
\min_G \max_D L(G, D)
\end{aligned}
\end{equation}

Considering the objective function in the Wasserstein GAN with gradient penalty~\cite{NIPS2017_7159}, we propose a loss function $L$ by adding an additional loss term for the generator. 
\begin{equation}
\begin{aligned}
L(G, & D, P)   = \E_{\xtilde \sim \PP_g} [D(\xtilde)+ \gamma  \abs{ P(G(\xtilde)) -  \hat{G}(\xtilde) } ] \\
    & - \E_{\vx \sim \PP_r} [D(\vx)]  
    + \lambda \E_{\vxh \sim \PP_{\vxh}} [( {\Vert \nabla_{\vxh} D(\vxh) \Vert}_2 - 1)^2] \\
\end{aligned}
\end{equation}
where $\PP_g$ and $\PP_r$ denote the generator distribution and the training data distribution. The authors in~\cite{NIPS2017_7159} implicitly defined $\PP_{\vxh}$ as sampling uniformly along straight lines between pairs of points sampled from $\PP_r$ and $\PP_g$. $\hat{G}$ is the layers in the generator $G$ before any locally connected layer. $\hat{G}$ is initialized by the parameters that are trained using~\cite{NIPS2017_7159}. $P$ is the transformation between the generator and the discriminator. $\gamma$ is the hyperparameter to weight between typical generator loss and the proposed additional generator loss. Gradient penalty coefficient $\lambda$ is 10 and $\gamma$ is 100 in all experiments. The learning rates for the discriminator and the generator are 0.0003 and 0.0001, respectively.

We believe that further studying on an optimization algorithm for non-typical images can improve the quality of generated results further. However, developing a novel optimization algorithm is beyond the scope of this paper.

%% file: sections/fig_locally_connect.tex
\begin{figure*}[!t] \begin{center}
\begin{minipage}{0.32\linewidth}
\centerline{\includegraphics[width=2in]{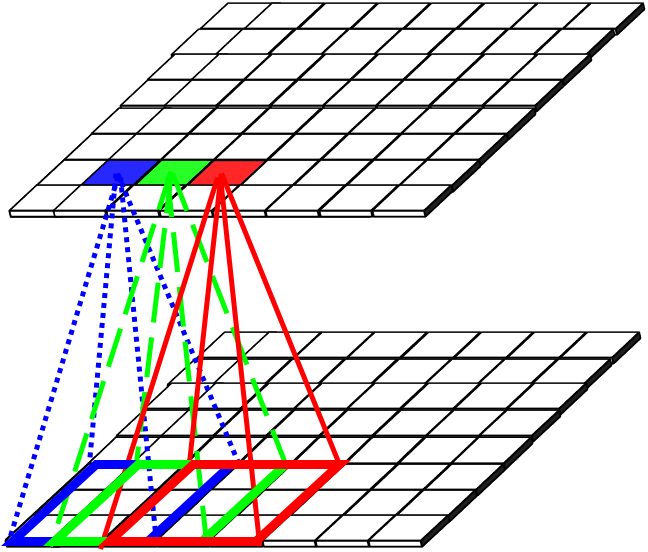}}
\centerline{\footnotesize (a) Convolution layer} 
\end{minipage}
\begin{minipage}{0.32\linewidth}
\centerline{\includegraphics[width=2in]{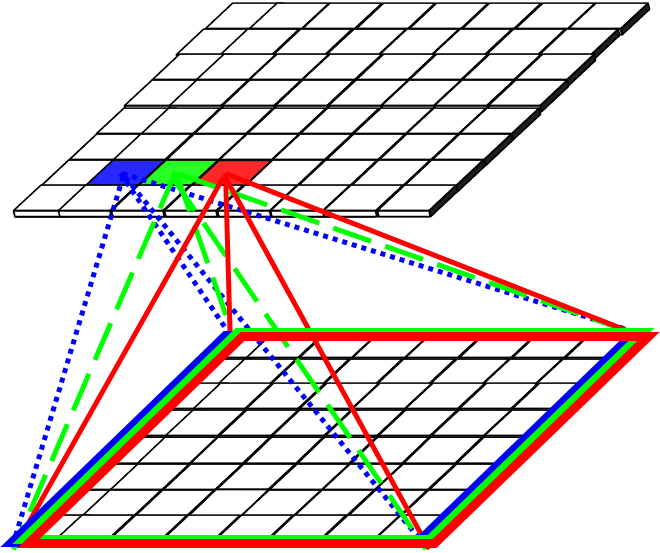}}
\centerline{\footnotesize (b) Fully connected layer} 
\end{minipage}
\begin{minipage}{0.32\linewidth}
\centerline{\includegraphics[width=2in]{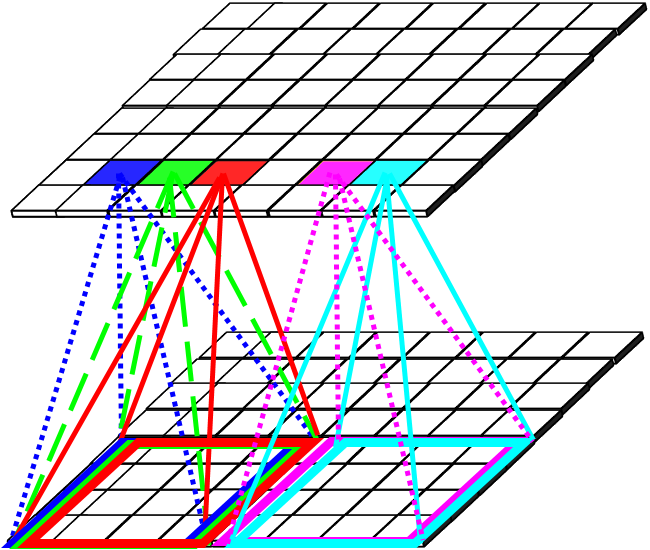}}
\centerline{\footnotesize (c) Locally connected layer} 
\end{minipage}
   \caption{Visual comparison of (a) convolution layer with filter-size of 3$\times$3, (b) fully connected layer, and (c) proposed locally connected layer with block-size of 4$\times$4. The locally connected layer operates comparable to block-based processing. Each region of output is produced by the summation of the multiplication of the corresponding region of input and shared weights. The weights are shared between blocks, but not between outputs in a block.}
\label{fig:locallyConnect}
\end{center}\end{figure*}

%% file: sections/results.tex
\graphicspath{ {images/} }

\section{Experiments and Results} \label{results}

\input{sections/tbl_result_cifar}

\subsection{Dataset} \label{dataset}
We experiment using the CIFAR-10 training dataset~\cite{cifar} and the LSUN bedroom training dataset~\cite{yu15lsun}. The CIFAR-10 dataset consists of 50,000 images with the resolution of 32$\times$32. The dataset includes images from 10 categories (airplane, automobile, bird, cat, deer, dog, frog, horse, ship, and truck). The LSUN bedroom dataset consists of 3,033,042 bedroom images. The images are scaled to 64$\times$64 following the previous works~\cite{NIPS2017_7159}. 

\subsection{Metric} \label{metric}
We use the Fr\'{e}chet Inception Distance (FID)~\cite{NIPS2017_7240} which was improved from the Inception score~\cite{Salimans} by considering the statistics of real data. The FID computes the Fr\'{e}chet distance (also known as Wasserstein-2 distance)~\cite{frechet1, frechet2} between the statistics of real data and that of generated samples. The distance is computed using the first two moments (mean and covariance) of activations from the last pooling layer in the Inception v3 model~\cite{InceptionV3}. The FID $d$ is computed as follows:
\begin{equation}
\begin{aligned}
d^2((\vm, \mC), (\vm_w, \mC_w)) =& ||\vm - \vm_w||_2^2 \quad + \\
& \text{Tr}(\mC + \mC_w - 2(\mC \mC_w)^{1/2})
\end{aligned}
\end{equation}
where $(\vm, \mC)$ and $(\vm_w, \mC_w)$ represent the mean and covariance of generated samples and of the real dataset. $(\vm_w, \mC_w)$ are measured using entire images in the training dataset. $(\vm, \mC)$ are computed using 50,000 generated images.

\subsection{Results}
We analyze the proposed method and the architectures in the TTUR method~\cite{NIPS2017_7240} by training them using the datasets in~\sref{dataset} and by evaluating them quantitatively and qualitatively. For quantitative comparison, we measure the Fr\'{e}chet Inception Distance (FID) in~\sref{metric}. \tref{tab:resultCIFAR} and \tref{tab:resultLSUN} show the quantitative results for the CIFAR-10 dataset~\cite{cifar} and the LSUN bedroom dataset~\cite{yu15lsun}, respectively. In both tables, we show the FIDs for three chroma subsampling ratios (4:4:4, 4:2:2, 4:2:0) and for four quality factors of quantization (100, 75, 50, 25). The first row shows the FID variations of real images by applying chroma subsampling and quantization. The second row presents the FID of the original TTUR method generating RGB images~\cite{NIPS2017_7240}. For this analysis, JPEG compression is processed as a post-processing step that follows the neural networks. The third row shows the result of the TTUR method for generating JPEG compressed images directly. In~\tref{tab:resultCIFAR}, the fourth row presents the result of the fully connected (FC) generator in the TTUR method for generating JPEG compressed images directly. In the last row, we show the result of the proposed method. 

We show visual results of the CIFAR-10 and the LSUN bedroom datasets in Figs.~\ref{fig:result_cifar} and~\ref{fig:result_lsun}, respectively. On the left side, we denote the quality factor for quantization and chroma subsampling ratios. We show the results of (100, 4:4:4), (100, 4:2:2), (100, 4:2:0), (75, 4:4:4), (50, 4:4:4), and (25, 4:4:4) from the first row to the last row. On the first column, we show the results of real images that are processed by the corresponding compression. The second column shows the results of the generated images using the original TTUR method~\cite{NIPS2017_7240}. The result images are first generated as RGB images and are then coded and decoded in a post-processing stage. The third column shows the result of the TTUR method~\cite{NIPS2017_7240}. In~\fref{fig:result_cifar}, the fourth column presents the result of the FC generator in the TTUR method~\cite{NIPS2017_7240}. The last column shows the results of the proposed method. The last three columns in~\fref{fig:result_cifar} and two columns in~\fref{fig:result_lsun} are the results of generating JPEG compressed images in the networks.

To compute the FID in the first row in both tables, we first encode and decode 50,000 training images and then compute the statistics of the processed images. We then estimate the FID between the statistics of the original training data and those of the processed images. For the CIFAR-10 dataset in~\tref{tab:resultCIFAR}, the distance is close to 0 using the chroma subsampling ratio of 4:4:4 and the quality factor of 100. It's quite small since the encoding/decoding does not impact the images much (only small rounding errors, etc). By decreasing quality factor and by subsampling from a larger region, the encoding/decoding distorts images more and hence, FID increases. Since decreasing quality factor by 25 affects images much more than adjusting chroma subsampling from 4:4:4 to 4:2:0, FID is also increased by a larger amount. 

\input{sections/tbl_result_lsun}

For the LSUN bedroom dataset in~\tref{tab:resultLSUN}, the distance of real data using the chroma subsampling ratio of 4:4:4 and the quality factor of 100 is much greater than that in the CIFAR-10 dataset. The FID is defined by the distance between the statistics of the entire training data and those of 50,000 processed or generated images. Since the number of images in the CIFAR-10 dataset is 50,000, the distance is quite small considering the encoding/decoding does not distort much. However, since the LSUN bedroom dataset contains 3,033,042 images, 50,000 images should be sampled to compute the statistics of the processed images. It causes a relatively larger FID for the LSUN bedroom dataset. It is also interesting to note that for the LSUN dataset, the quality factor of 75 is better than that of 100 in most experiments. We believe since the bedroom images often have continuous tone because of its contents or pre-processing, discarding high-frequency components decreases FID.

FIDs in the second row is computed by first generating RGB images using the TTUR method~\cite{NIPS2017_7240} and by encoding/decoding the generated RGB images. As generating RGB images have been studied a lot in recent years and the TTUR method is one of the state-of-the-art methods, generated RGB images are quite visually plausible. FID is increased by encoding/decoding the images using a lower quality factor and subsampling from a larger region. Some of the distortions can be visually observed in Figs.~\ref{fig:result_cifar} and~\ref{fig:result_lsun}. 

The third row presents applying the same method to generate JPEG encoded images. The results demonstrate that directly applying the method does not produce competitive results. The fourth row in~\tref{tab:resultCIFAR} shows the results of applying the FC generator in the TTUR method~\cite{NIPS2017_7240} for generating encoded images. While FC generator often performs poorer than the selected TTUR method for generating typical images, we tried the FC generator to avoid extensively applied convolution layers in the TTUR method. However, the FC generator does not perform well even for generating encoded images. 

The last row in both tables shows the results of the proposed method. The proposed method achieves promising results for generating JPEG encoded images directly. The proposed method outperforms applying the TTUR method for generating JPEG encoded image directly. Moreover, the proposed method is competitive to the method that generates RGB images using the TTUR method and compresses them by post-processing. 

\input{sections/fig_result_cifar}
\input{sections/fig_result_lsun}

%% file: sections/tbl_result_cifar.tex
\begin{table*}
\begin{center}
\caption{Quantitative comparison using FIDs for the CIFAR-10 dataset~\cite{cifar}. The lower FID means the better result. The first and second rows show FIDs of training images and of generated RGB images using the TTUR method~\cite{NIPS2017_7240} by processing compression in a post-processing stage. The last three rows present generating compressed data directly using the TTUR method~\cite{NIPS2017_7240}, the FC Generator in the TTUR~\cite{NIPS2017_7240}, and the proposed method.} 
\label{tab:resultCIFAR}
\renewcommand{\arraystretch}{1.2} 
\begin{tabu}{X[c]|X[c]|X[c]|X[c]|X[c]|X[c]|X[c]} 
\hline
\multirow{2}{*}{Method} & Generator  & Chroma & \multicolumn{4}{c}{Quality factor} \\
\cline{4-7}
		 & output & subsampling & 100 & 75 & 50 & 25 \\
\hline\hline
            & 	& 4:4:4 & 0.02 & 6.49 & 16.89 & 35.76 \\
Real data	& - & 4:2:2 & 0.68 & 10.59 & 23.89 & 45.40 \\
	 		& 	& 4:2:0 & 1.86 & 16.30 & 32.47 & 55.83 \\
\hline
\multirow{3}{*}{TTUR~\cite{NIPS2017_7240}} 	& 		& 4:4:4 & 26.10 & \textbf{31.77} & 41.02 & 54.21 \\
 			& RGB image 	& 4:2:2 & \textbf{25.79} & 37.03 & 48.45 & 63.37 \\
 			& 		& 4:2:0 & 26.90 & \textbf{43.31} & \textbf{56.22} & 72.30 \\
\hline
\multirow{3}{*}{TTUR~\cite{NIPS2017_7240}}	& & 4:4:4 & 68.80 & 75.23 & 82.20 & 89.80 \\
						& & 4:2:2 & 62.23 & 71.85 & 81.30 & 91.39 \\
						& & 4:2:0 & 75.99 & 93.07 & 102.39 & 109.08 \\
\cline{1-1}\cline{3-7}
\multirow{2}{*}{FC generator}		& JPEG	& 4:4:4 & 83.31 & 87.71 & 94.40 & 96.67 \\
\multirow{2}{*}{\cite{NIPS2017_7240}} 	& compressed & 4:2:2 & 91.32 &104.28 & 110.62 & 108.09 \\
						& image	& 4:2:0 & 71.37 & 93.27 & 101.73 & 104.64 \\
\cline{1-1}\cline{3-7}
\multirow{2}{*}{Proposed} & & 4:4:4 & \textbf{25.57} & 31.78 & \textbf{40.94} & \textbf{54.20} \\
\multirow{2}{*}{method}	& & 4:2:2 & 25.80 & \textbf{37.01} & \textbf{48.25} & \textbf{63.22} \\
 						& & 4:2:0 & \textbf{26.78} & 43.35 & 56.30 & \textbf{72.19} \\
\hline
\end{tabu}
\end{center}
\end{table*}

%% file: sections/tbl_result_lsun.tex
\begin{table*}
\begin{center}
\caption{Quantitative comparison using FIDs for the LSUN dataset~\cite{yu15lsun}. The lower FID means the better result. The first and second rows show FIDs of training images and of generated RGB images using the TTUR method~\cite{NIPS2017_7240} by processing compression in a post-processing stage. The last two rows present generating compressed data directly using the TTUR~\cite{NIPS2017_7240} and the proposed method.}
\label{tab:resultLSUN}
\renewcommand{\arraystretch}{1.2} 
\begin{tabu}{X[c]|X[c]|X[c]|X[c]|X[c]|X[c]|X[c]} 
\hline
\multirow{2}{*}{Method} & Generator  & Chroma & \multicolumn{4}{c}{Quality factor} \\
\cline{4-7}
		 & output & subsampling & 100 & 75 & 50 & 25 \\
\hline\hline
            & 	& 4:4:4 & 9.96 & 6.82 & 7.57 & 18.26 \\
Real data	& - & 4:2:2 & 10.36 & 7.13 & 8.18 & 18.94 \\
	 		& 	& 4:2:0 & 11.37 & 10.22 & 12.31 & 23.01 \\
\hline
\multirow{3}{*}{TTUR~\cite{NIPS2017_7240}}  & 	& 4:4:4 & 12.44 & 10.24 & 11.35 & \textbf{23.17} \\
 				& RGB image 	& 4:2:2 & 13.41 & 12.05 & 13.34 & 24.46 \\
 				& 		& 4:2:0 & 14.80 & 16.00 & 17.98 & 27.99 \\
\hline
\multirow{3}{*}{TTUR~\cite{NIPS2017_7240}}	& & 4:4:4 & 35.70 & 27.78 & 30.70 & 45.39 \\
			& \multirow{2}{*}{JPEG} 		& 4:2:2 & 43.35 & 34.51 & 36.19 & 50.22 \\
			& \multirow{2}{*}{compressed} 	& 4:2:0 & 71.32 & 62.18 & 67.43 & 75.30 \\
\cline{1-1}\cline{3-7}
\multirow{2}{*}{Proposed}	& \multirow{2}{*}{image} & 4:4:4 & \textbf{12.13} & \textbf{9.99} & \textbf{11.25} & 23.18 \\
\multirow{2}{*}{method}		& 	& 4:2:2 & \textbf{12.96} & \textbf{11.57} & \textbf{13.12} & \textbf{24.28} \\
 							& 	& 4:2:0 & \textbf{14.21} & \textbf{15.41} & \textbf{17.63} & \textbf{27.82} \\
\hline
\end{tabu}
\end{center}
\end{table*}

%% file: sections/fig_result_cifar.tex
\begin{figure*}[!t] \begin{center}
\begin{minipage}{0.05\linewidth}
100, 4:4:4
\end{minipage}
\begin{minipage}{0.18\linewidth}
\centerline{\includegraphics[scale=1.46]{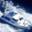}\includegraphics[scale=1.46]{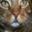}}
\centerline{\includegraphics[scale=1.46]{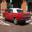}\includegraphics[scale=1.46]{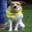}}
\end{minipage}
\begin{minipage}{0.18\linewidth}
\centerline{\includegraphics[scale=1.46]{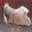}\includegraphics[scale=1.46]{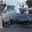}}
\centerline{\includegraphics[scale=1.46]{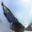}\includegraphics[scale=1.46]{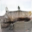}}
\end{minipage}
\begin{minipage}{0.18\linewidth}
\centerline{\includegraphics[scale=1.46]{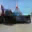}\includegraphics[scale=1.46]{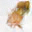}}
\centerline{\includegraphics[scale=1.46]{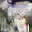}\includegraphics[scale=1.46]{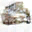}}
\end{minipage}
\begin{minipage}{0.18\linewidth}
\centerline{\includegraphics[scale=1.46]{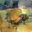}\includegraphics[scale=1.46]{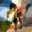}}
\centerline{\includegraphics[scale=1.46]{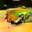}\includegraphics[scale=1.46]{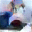}}
\end{minipage}
\begin{minipage}{0.18\linewidth}
\centerline{\includegraphics[scale=1.46]{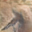}\includegraphics[scale=1.46]{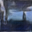}}
\centerline{\includegraphics[scale=1.46]{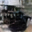}\includegraphics[scale=1.46]{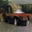}}
\end{minipage}
\\
\vspace{0.3cm}
\begin{minipage}{0.05\linewidth}
100, 4:2:2
\end{minipage}
\begin{minipage}{0.18\linewidth}
\centerline{\includegraphics[scale=1.46]{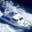}\includegraphics[scale=1.46]{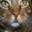}}
\centerline{\includegraphics[scale=1.46]{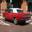}\includegraphics[scale=1.46]{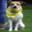}}
\end{minipage}
\begin{minipage}{0.18\linewidth}
\centerline{\includegraphics[scale=1.46]{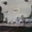}\includegraphics[scale=1.46]{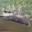}}
\centerline{\includegraphics[scale=1.46]{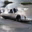}\includegraphics[scale=1.46]{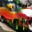}}
\end{minipage}
\begin{minipage}{0.18\linewidth}
\centerline{\includegraphics[scale=1.46]{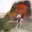}\includegraphics[scale=1.46]{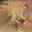}}
\centerline{\includegraphics[scale=1.46]{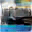}\includegraphics[scale=1.46]{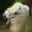}}
\end{minipage}
\begin{minipage}{0.18\linewidth}
\centerline{\includegraphics[scale=1.46]{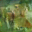}\includegraphics[scale=1.46]{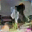}}
\centerline{\includegraphics[scale=1.46]{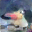}\includegraphics[scale=1.46]{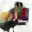}}
\end{minipage}
\begin{minipage}{0.18\linewidth}
\centerline{\includegraphics[scale=1.46]{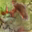}\includegraphics[scale=1.46]{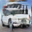}}
\centerline{\includegraphics[scale=1.46]{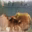}\includegraphics[scale=1.46]{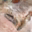}}
\end{minipage}
\\
\vspace{0.3cm}
\begin{minipage}{0.05\linewidth}
100, 4:2:0
\end{minipage}
\begin{minipage}{0.18\linewidth}
\centerline{\includegraphics[scale=1.46]{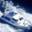}\includegraphics[scale=1.46]{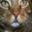}}
\centerline{\includegraphics[scale=1.46]{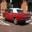}\includegraphics[scale=1.46]{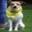}}
\end{minipage}
\begin{minipage}{0.18\linewidth}
\centerline{\includegraphics[scale=1.46]{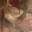}\includegraphics[scale=1.46]{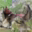}}
\centerline{\includegraphics[scale=1.46]{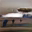}\includegraphics[scale=1.46]{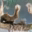}}
\end{minipage}
\begin{minipage}{0.18\linewidth}
\centerline{\includegraphics[scale=1.46]{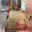}\includegraphics[scale=1.46]{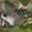}}
\centerline{\includegraphics[scale=1.46]{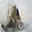}\includegraphics[scale=1.46]{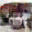}}
\end{minipage}
\begin{minipage}{0.18\linewidth}
\centerline{\includegraphics[scale=1.46]{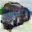}\includegraphics[scale=1.46]{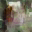}}
\centerline{\includegraphics[scale=1.46]{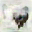}\includegraphics[scale=1.46]{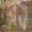}}
\end{minipage}
\begin{minipage}{0.18\linewidth}
\centerline{\includegraphics[scale=1.46]{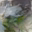}\includegraphics[scale=1.46]{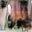}}
\centerline{\includegraphics[scale=1.46]{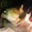}\includegraphics[scale=1.46]{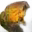}}
\end{minipage}
\\
\vspace{0.3cm}
\begin{minipage}{0.05\linewidth}
75, 4:4:4
\end{minipage}
\begin{minipage}{0.18\linewidth}
\centerline{\includegraphics[scale=1.46]{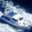}\includegraphics[scale=1.46]{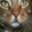}}
\centerline{\includegraphics[scale=1.46]{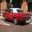}\includegraphics[scale=1.46]{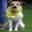}}
\end{minipage}
\begin{minipage}{0.18\linewidth}
\centerline{\includegraphics[scale=1.46]{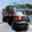}\includegraphics[scale=1.46]{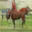}}
\centerline{\includegraphics[scale=1.46]{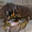}\includegraphics[scale=1.46]{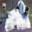}}
\end{minipage}
\begin{minipage}{0.18\linewidth}
\centerline{\includegraphics[scale=1.46]{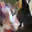}\includegraphics[scale=1.46]{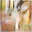}}
\centerline{\includegraphics[scale=1.46]{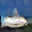}\includegraphics[scale=1.46]{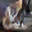}}
\end{minipage}
\begin{minipage}{0.18\linewidth}
\centerline{\includegraphics[scale=1.46]{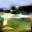}\includegraphics[scale=1.46]{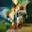}}
\centerline{\includegraphics[scale=1.46]{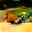}\includegraphics[scale=1.46]{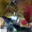}}
\end{minipage}
\begin{minipage}{0.18\linewidth}
\centerline{\includegraphics[scale=1.46]{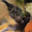}\includegraphics[scale=1.46]{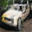}}
\centerline{\includegraphics[scale=1.46]{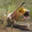}\includegraphics[scale=1.46]{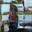}}
\end{minipage}
\\
\vspace{0.3cm}
\begin{minipage}{0.05\linewidth}
50, 4:4:4
\end{minipage}
\begin{minipage}{0.18\linewidth}
\centerline{\includegraphics[scale=1.46]{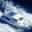}\includegraphics[scale=1.46]{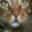}}
\centerline{\includegraphics[scale=1.46]{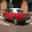}\includegraphics[scale=1.46]{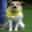}}
\end{minipage}
\begin{minipage}{0.18\linewidth}
\centerline{\includegraphics[scale=1.46]{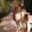}\includegraphics[scale=1.46]{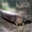}}
\centerline{\includegraphics[scale=1.46]{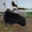}\includegraphics[scale=1.46]{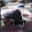}}
\end{minipage}
\begin{minipage}{0.18\linewidth}
\centerline{\includegraphics[scale=1.46]{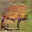}\includegraphics[scale=1.46]{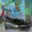}}
\centerline{\includegraphics[scale=1.46]{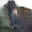}\includegraphics[scale=1.46]{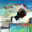}}
\end{minipage}
\begin{minipage}{0.18\linewidth}
\centerline{\includegraphics[scale=1.46]{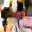}\includegraphics[scale=1.46]{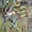}}
\centerline{\includegraphics[scale=1.46]{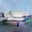}\includegraphics[scale=1.46]{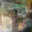}}
\end{minipage}
\begin{minipage}{0.18\linewidth}
\centerline{\includegraphics[scale=1.46]{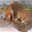}\includegraphics[scale=1.46]{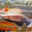}}
\centerline{\includegraphics[scale=1.46]{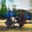}\includegraphics[scale=1.46]{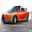}}
\end{minipage}
\\
\vspace{0.3cm}
\begin{minipage}{0.05\linewidth}
25, 4:4:4
\end{minipage}
\begin{minipage}{0.18\linewidth}
\centerline{\includegraphics[scale=1.46]{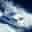}\includegraphics[scale=1.46]{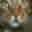}}
\centerline{\includegraphics[scale=1.46]{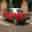}\includegraphics[scale=1.46]{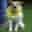}}
\centerline{\footnotesize (a) Real data} 
\end{minipage}
\begin{minipage}{0.18\linewidth}
\centerline{\includegraphics[scale=1.46]{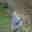}\includegraphics[scale=1.46]{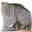}}
\centerline{\includegraphics[scale=1.46]{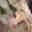}\includegraphics[scale=1.46]{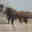}}
\centerline{\footnotesize (b) TTUR (RGB)} 
\end{minipage}
\begin{minipage}{0.18\linewidth}
\centerline{\includegraphics[scale=1.46]{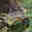}\includegraphics[scale=1.46]{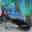}}
\centerline{\includegraphics[scale=1.46]{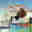}\includegraphics[scale=1.46]{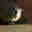}}
\centerline{\footnotesize (c) TTUR (encoded)} 
\end{minipage}
\begin{minipage}{0.18\linewidth}
\centerline{\includegraphics[scale=1.46]{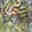}\includegraphics[scale=1.46]{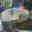}}
\centerline{\includegraphics[scale=1.46]{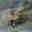}\includegraphics[scale=1.46]{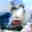}}
\centerline{\footnotesize (d) FC generator (encoded)} 
\end{minipage}
\begin{minipage}{0.18\linewidth}
\centerline{\includegraphics[scale=1.46]{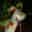}\includegraphics[scale=1.46]{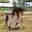}}
\centerline{\includegraphics[scale=1.46]{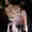}\includegraphics[scale=1.46]{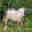}}
\centerline{\footnotesize (e) Proposed method} 
\end{minipage}
   \caption{Visual results of generating compressed images using the CIFAR-10 dataset~\cite{cifar}. On the left side, we denote the quality factor for quantization and chroma subsampling ratios. We show the results of (100, 4:4:4), (100, 4:2:2), (100, 4:2:0), (75, 4:4:4), (50, 4:4:4), and (25, 4:4:4) from the first row to the last row. The first and second columns show the results of real images and the original TTUR method~\cite{NIPS2017_7240} that are processed by the corresponding encoding/decoding. The third and fourth columns show the results of the TTUR method and the FC generator in the TTUR method that generates compressed data directly. The last column shows the result of the proposed method.}
\label{fig:result_cifar}
\end{center}\end{figure*}

%% file: sections/fig_result_lsun.tex
\begin{figure*}[!t] \begin{center}
\begin{minipage}{0.05\linewidth}
100, 4:4:4
\end{minipage}
\begin{minipage}{0.22\linewidth}
\centerline{\includegraphics[scale=0.73]{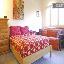}\includegraphics[scale=0.73]{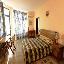}}
\centerline{\includegraphics[scale=0.73]{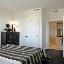}\includegraphics[scale=0.73]{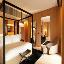}}
\end{minipage}
\begin{minipage}{0.22\linewidth}
\centerline{\includegraphics[scale=0.73]{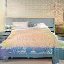}\includegraphics[scale=0.73]{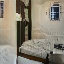}}
\centerline{\includegraphics[scale=0.73]{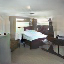}\includegraphics[scale=0.73]{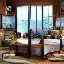}}
\end{minipage}
\begin{minipage}{0.22\linewidth}
\centerline{\includegraphics[scale=0.73]{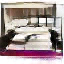}\includegraphics[scale=0.73]{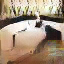}}
\centerline{\includegraphics[scale=0.73]{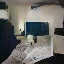}\includegraphics[scale=0.73]{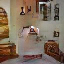}}
\end{minipage}
\begin{minipage}{0.22\linewidth}
\centerline{\includegraphics[scale=0.73]{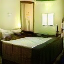}\includegraphics[scale=0.73]{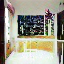}}
\centerline{\includegraphics[scale=0.73]{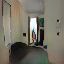}\includegraphics[scale=0.73]{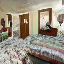}}
\end{minipage}
\\
\vspace{0.3cm}
\begin{minipage}{0.05\linewidth}
100, 4:2:2
\end{minipage}
\begin{minipage}{0.22\linewidth}
\centerline{\includegraphics[scale=0.73]{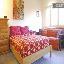}\includegraphics[scale=0.73]{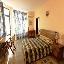}}
\centerline{\includegraphics[scale=0.73]{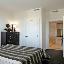}\includegraphics[scale=0.73]{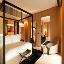}}
\end{minipage}
\begin{minipage}{0.22\linewidth}
\centerline{\includegraphics[scale=0.73]{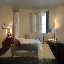}\includegraphics[scale=0.73]{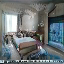}}
\centerline{\includegraphics[scale=0.73]{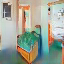}\includegraphics[scale=0.73]{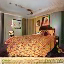}}
\end{minipage}
\begin{minipage}{0.22\linewidth}
\centerline{\includegraphics[scale=0.73]{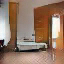}\includegraphics[scale=0.73]{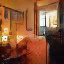}}
\centerline{\includegraphics[scale=0.73]{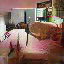}\includegraphics[scale=0.73]{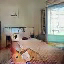}}
\end{minipage}
\begin{minipage}{0.22\linewidth}
\centerline{\includegraphics[scale=0.73]{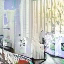}\includegraphics[scale=0.73]{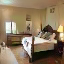}}
\centerline{\includegraphics[scale=0.73]{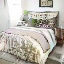}\includegraphics[scale=0.73]{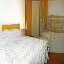}}
\end{minipage}
\\
\vspace{0.3cm}
\begin{minipage}{0.05\linewidth}
100, 4:2:0
\end{minipage}
\begin{minipage}{0.22\linewidth}
\centerline{\includegraphics[scale=0.73]{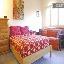}\includegraphics[scale=0.73]{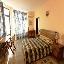}}
\centerline{\includegraphics[scale=0.73]{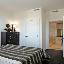}\includegraphics[scale=0.73]{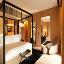}}
\end{minipage}
\begin{minipage}{0.22\linewidth}
\centerline{\includegraphics[scale=0.73]{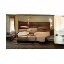}\includegraphics[scale=0.73]{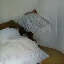}}
\centerline{\includegraphics[scale=0.73]{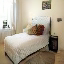}\includegraphics[scale=0.73]{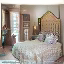}}
\end{minipage}
\begin{minipage}{0.22\linewidth}
\centerline{\includegraphics[scale=0.73]{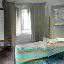}\includegraphics[scale=0.73]{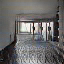}}
\centerline{\includegraphics[scale=0.73]{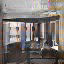}\includegraphics[scale=0.73]{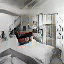}}
\end{minipage}
\begin{minipage}{0.22\linewidth}
\centerline{\includegraphics[scale=0.73]{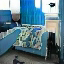}\includegraphics[scale=0.73]{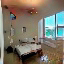}}
\centerline{\includegraphics[scale=0.73]{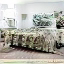}\includegraphics[scale=0.73]{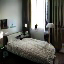}}
\end{minipage}
\\
\vspace{0.3cm}
\begin{minipage}{0.05\linewidth}
75, 4:4:4
\end{minipage}
\begin{minipage}{0.22\linewidth}
\centerline{\includegraphics[scale=0.73]{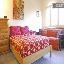}\includegraphics[scale=0.73]{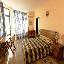}}
\centerline{\includegraphics[scale=0.73]{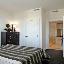}\includegraphics[scale=0.73]{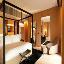}}
\end{minipage}
\begin{minipage}{0.22\linewidth}
\centerline{\includegraphics[scale=0.73]{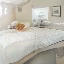}\includegraphics[scale=0.73]{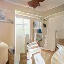}}
\centerline{\includegraphics[scale=0.73]{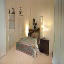}\includegraphics[scale=0.73]{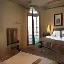}}
\end{minipage}
\begin{minipage}{0.22\linewidth}
\centerline{\includegraphics[scale=0.73]{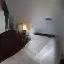}\includegraphics[scale=0.73]{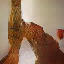}}
\centerline{\includegraphics[scale=0.73]{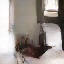}\includegraphics[scale=0.73]{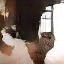}}
\end{minipage}
\begin{minipage}{0.22\linewidth}
\centerline{\includegraphics[scale=0.73]{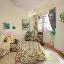}\includegraphics[scale=0.73]{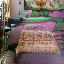}}
\centerline{\includegraphics[scale=0.73]{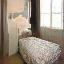}\includegraphics[scale=0.73]{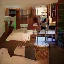}}
\end{minipage}
\\
\vspace{0.3cm}
\begin{minipage}{0.05\linewidth}
50, 4:4:4
\end{minipage}
\begin{minipage}{0.22\linewidth}
\centerline{\includegraphics[scale=0.73]{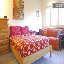}\includegraphics[scale=0.73]{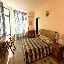}}
\centerline{\includegraphics[scale=0.73]{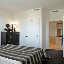}\includegraphics[scale=0.73]{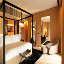}}
\end{minipage}
\begin{minipage}{0.22\linewidth}
\centerline{\includegraphics[scale=0.73]{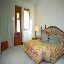}\includegraphics[scale=0.73]{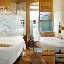}}
\centerline{\includegraphics[scale=0.73]{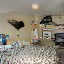}\includegraphics[scale=0.73]{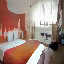}}
\end{minipage}
\begin{minipage}{0.22\linewidth}
\centerline{\includegraphics[scale=0.73]{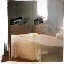}\includegraphics[scale=0.73]{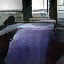}}
\centerline{\includegraphics[scale=0.73]{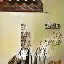}\includegraphics[scale=0.73]{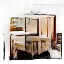}}
\end{minipage}
\begin{minipage}{0.22\linewidth}
\centerline{\includegraphics[scale=0.73]{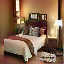}\includegraphics[scale=0.73]{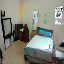}}
\centerline{\includegraphics[scale=0.73]{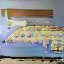}\includegraphics[scale=0.73]{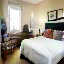}}
\end{minipage}
\\
\vspace{0.3cm}
\begin{minipage}{0.05\linewidth}
25, 4:4:4
\end{minipage}
\begin{minipage}{0.22\linewidth}
\centerline{\includegraphics[scale=0.73]{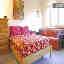}\includegraphics[scale=0.73]{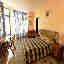}}
\centerline{\includegraphics[scale=0.73]{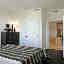}\includegraphics[scale=0.73]{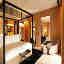}}
\centerline{\footnotesize (a) Real data} 
\end{minipage}
\begin{minipage}{0.22\linewidth}
\centerline{\includegraphics[scale=0.73]{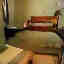}\includegraphics[scale=0.73]{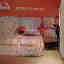}}
\centerline{\includegraphics[scale=0.73]{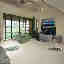}\includegraphics[scale=0.73]{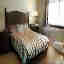}}
\centerline{\footnotesize (b) TTUR (RGB)} 
\end{minipage}
\begin{minipage}{0.22\linewidth}
\centerline{\includegraphics[scale=0.73]{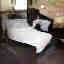}\includegraphics[scale=0.73]{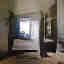}}
\centerline{\includegraphics[scale=0.73]{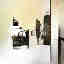}\includegraphics[scale=0.73]{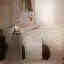}}
\centerline{\footnotesize (c) TTUR (encoded)} 
\end{minipage}
\begin{minipage}{0.22\linewidth}
\centerline{\includegraphics[scale=0.73]{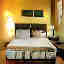}\includegraphics[scale=0.73]{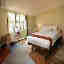}}
\centerline{\includegraphics[scale=0.73]{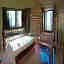}\includegraphics[scale=0.73]{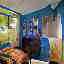}}
\centerline{\footnotesize (d) Proposed method} 
\end{minipage}
   \caption{Visual results of generating compressed images using the LSUN dataset~\cite{yu15lsun}. On the left side, we denote the quality factor for quantization and chroma subsampling ratios. We show the results of (100, 4:4:4), (100, 4:2:2), (100, 4:2:0), (75, 4:4:4), (50, 4:4:4), and (25, 4:4:4) from the first row to the last row. The first and second columns show the results of real images and the original TTUR method~\cite{NIPS2017_7240} that are processed by the corresponding encoding/decoding. The third column presents the results of the TTUR method that generates compressed data directly. The last column shows the result of the proposed method.}
\label{fig:result_lsun}
\end{center}\end{figure*}

%% file: sections/conclusions.tex
\section{Conclusion} \label{conclusions}
We present a generative adversarial network framework that combines image generation and compression by generating compressed images directly. We propose a novel generator consisting of the proposed locally connected layers, chroma subsampling layers, quantization layers, residual blocks, and convolution layers. We also present training strategies for the proposed framework including the loss function and the transformation between the generator and the discriminator. We demonstrate that the proposed framework outperforms applying the state-of-the-art GANs for generating compressed data directly. Moreover, we show that the proposed method achieves competitive results comparing to generating raw RGB images using one of the state-of-the-art methods and compressing the images by post-processing. We believe that the proposed method can be further improved by investigating optimization algorithms for learning to generate compressed data. We also consider the scenario where the proposed method can serve as a baseline method for further studies.